\documentclass[11pt,a4paper]{article}
\usepackage[margin=1in]{geometry}
\usepackage{graphicx}
\usepackage{amsmath}
\usepackage{amssymb}
\usepackage{booktabs}
\usepackage[hypertexnames=false]{hyperref}
\usepackage{xcolor}
\usepackage{fancyhdr}
\usepackage{float}
\usepackage{natbib}

\title{\textbf{Causely: A Causal Intelligence Layer Reliability Workflows\\
Benchmark Study on AI SRE and Reliability Workflows}}
\author{Causely}
\date{\today}

\begin{document}

% Custom title page
\begin{titlepage}
  \centering
  \vspace*{1.5cm}
  \includegraphics[height=1.2cm]{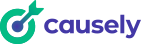}

  \vspace{2.5cm}
  {\LARGE \textbf{Causely: A Causal Intelligence Layer for Enterprise AI}}

  \vspace{1.5cm}
  {\large Benchmark Study on SRE and Reliability Workflows}

  \vspace{2cm}

  {\normalsize
    Dhairya Dalal \quad
    Endre Sara \quad
    Ben Yemini \quad
    Christine Miller \quad
    Shmuel Kliger
  }

  \vspace{0.4cm}

  {\small
    \texttt{ddalal@causely.ai},
    \texttt{esara@causely.ai},
    \texttt{byemini@causely.ai},
    \texttt{cmiller@causely.ai},
    \texttt{skliger@causely.ai}
  }

  \vspace{0.8cm}
  {\normalsize \today}

  \vfill
  {\small \textit{White Paper}}
\end{titlepage}

% Abstract
\begin{abstract}
AI agents deployed into SRE workflows currently derive their understanding of environment state from raw observability telemetry at query time, paying a semantic-interpretation tax in tokens, latency, and inferential reliability. We propose Causely, a causal intelligence layer that maintains a structured representation of environment topology, attribute dependencies, and causal relationships that are anchroed to a ontological representation of the managed environment. Causely transforms raw telemetry into a live, queryable model providing the semantic and causal foundation AI agents require to diagnose, evaluate impact, and act safely in production. We evaluate this value proposition through a benchmark study conducted in a controlled setting with injected faults in a 24-microservice OpenTelemetry demo application. Our experiments compare four agent configurations (Claude Code, OpenAI Codex, HolmesGPT with Sonnet and Gemini backends). Experiments are run with and without access to Causely under two scenarios: an active incident and a healthy baseline. On the active-fault scenario, causal grounding reduces mean time-to-diagnosis by 63\%, mean token consumption by 60\%, and mean tool-call count by 78\%, compressing the investigation footprint by 4.8$\times$ and lowering direct API cost per run by 57\%; root-cause-diagnosis accuracy rises from 75\% to 100\%. 
\end{abstract}

\newpage
\tableofcontents
\newpage

%==============================================================================
% SECTION 1: OVERVIEW
%==============================================================================

\section{Introduction}
\label{sec:intro}

\subsection{Rise of Agentic AI and the Promise of Continuous Reliability}

Modern software systems are growing in complexity faster than engineering teams can manage through individual effort. Accelerated delivery cycles, complex microservices architectures with cross-service dependencies, and the widespread adoption of AI-generated code have dramatically increased the rate at which new components, dependencies, and failure modes manifest in production environments. McKinsey's 2025 State of AI report found that 78\% of organizations have deployed AI across at least one business function, with C-suite leaders increasingly identifying agentic AI adoption as a competitive imperative \cite{mckinsey2025stateofai,mckinsey2025workplace}. Engineering and platform teams have responded by integrating agentic AI into continuous reliability, the practice of using observability data throughout the DevOps lifecycle to proactively maintain system health across pre-production and production environments, with applications spanning incident detection, root cause analysis, change safety evaluation, and capacity planning \cite{devops2024continuous,datadog2024feedback,computerworld2024observability}. Engineering teams have begun exploring both vendor-supplied and in-house AI solutions for reliability workflows, ranging from agents that assist engineers in investigating and managing alerts to fully autonomous systems capable of end-to-end incident triage and remediation \cite{gartner2024srehicycle,gartner2025aisre}. Yet as these deployments remain in their early stages, the effectiveness of AI agents for continuous reliability has not been rigorously quantified, and whether their performance in practice matches the promise attributed to them remains an open empirical question.

\subsection{The Telemetry-to-Knowledge Gap for AI Agents}

An AI agent answering SRE questions must first understand the state of the environment it is reasoning about: what services exist, how they depend on one another, and what has deviated from normal. Observability platforms are the starting point for this understanding, supplying log streams, latency and error-rates, distributed traces, and infrastructure event feeds \cite{beyer2016sre}. However, observability on face is insufficient to describe environment state: the telemetry must be interpreted and contextualized. Error patterns must be identified in logs, deviations measured against baselines, activity correlated across service boundaries, and events ordered into coherent narratives before any signal becomes knowledge the agent can act on.

Without a persistent context layer, AI agents perform this interpretation on every query, pulling raw telemetry into the context window and deriving a semantic understanding of environment state within a single inference loop. That understanding is rarely clean. Topologies are dynamic, deployments roll over, pods and containers churn, and the agent must piece together services, dependencies, and causal relationships. Entering an investigation with limited context, the agent retraces logs, alerts, and anomalies after the fact. If the environmentstate has shifted, the agent's conclusions aregrounded in stale evidence, a problem compounded by degradation of model performance over long contexts \cite{liu2024lostmiddle,goyal2025contextlength}. As a result, each subsequent agent query costs requires more tool calls and telemetry to be pulled into the context window, making the process expensive as well as fragile.

\subsection{The Hidden Cost of Environment Interpretation}

The token cost of this interpretation work is largely invisible to enterprises today. Most AI deployments are purchased through flat-rate seat subscriptions that bundle model compute at a fixed monthly price, so an agent that consumes 100K tokens per investigation incurs no more visible marginal cost than one that consumes 10K \cite{anthropic2025pricing}. The true token economics of deriving environment understanding from raw telemetry are rendered invisible: the most inefficient agents, under current billing, look identical to the most efficient ones.

However, the compute economics are posing significant threats to the feasibility of modern AI providers. Financial disclosures reported by Fortune project OpenAI's operating losses to exceed \$74 billion in 2028, driven primarily by compute spending \cite{fortune2025openai}, and Anthropic began moving enterprise customers from seat-based subscriptions to per-token usage billing in late 2025, a change reported to potentially triple costs for heavy consumers \cite{theinformation2025anthropic}. As provider pricing converges on unit economics aligned with actual compute, the semantic-interpretation tax described above will shift from a subsidized inefficiency into a directly billable line item, and the agents that spend the most tokens deriving environment understanding will become financial liabilities on enterprise IT budgets.

\subsection{Motivation}

Addressing the limitations outlined above requires transforming raw observability data into structured, queryable knowledge that AI agents can consume directly. To this end, we propose the Causely \textbf{causal intelligence layer}: a persistent, continuously maintained representation of the environment's topology, attribute dependencies, live telemetry, and causal dynamics, defined formally in Section~\ref{sec:ci}. Causely is a proprietary implementation of this layer for managed cloud-native environments. The motivation of this study is to evaluate empirically whether supplying AI agents with a causal intelligence layer, instantiated via Causely, materially improves their performance on common continuous reliability use cases relative to operating directly on raw telemetry.

\subsection{Study Overview}

This paper presents a controlled empirical benchmark evaluating whether access to a causal intelligence layer improves AI agent performance on common continuous reliability tasks. We frame the study around a single broad research question and decompose it into three sub-questions aligned with the principal dimensions along which AI agents operating over observability data are evaluated in practice:
\begin{itemize}
  \item[\textbf{RQ.}] Does a causal intelligence layer materially improve the performance of AI agents on common SRE tasks?
  \begin{itemize}
    \item[\textbf{RQ.1}] \textit{(Latency)} How does access to a causal intelligence layer impact end-to-end time from query submission to diagnosis?
    \item[\textbf{RQ.2}] \textit{(Accuracy)} How does access to a causal intelligence layer impact diagnostic correctness, both under active faults and under healthy contexts?
    \item[\textbf{RQ.3}] \textit{(Token consumption)} How does access to a causal intelligence layer impact the volume of tokens required by the agent to support a reliability task?
  \end{itemize}
\end{itemize}

To evaluate these questions, we designed a benchmark comprising 72 experimental runs spanning four AI agent configurations (two productivity AI agents and two purpose-built Ops AI agents), two operational scenarios (a healthy baseline and an active fault), and two conditions (with and without access to Causely).

%==============================================================================
% SECTION 2: CAUSAL INTELLIGENCE
%==============================================================================

\section{Causal Intelligence}
\label{sec:ci}

\subsection{The Causal Intelligence Layer}

We present Causely, a \textbf{causal intelligence layer} which provides a dynamically updated, semantically structured, and causality-contextualized representation of environment state, exposed to AI agents as structured context. Formally, we define the causal intelligence layer as
\[
\mathcal{CI} = (\mathcal{G}_T,\, \mathcal{K}_C,\, \mathcal{G}_C,\, \mathcal{G}_A),
\]
where $\mathcal{G}_T$ is the \textit{topology graph} capturing the managed entities in an environment, $\mathcal{K}_C$ is the \textit{causal knowledge base} encoding global domain knowledge about common failure modes and their symptom closures, $\mathcal{G}_C$ is the \textit{causality graph} expressing the environment-specific probabilistic cause-and-effect structure, and $\mathcal{G}_A$ is the \textit{attribute dependency graph} describing functional relationships between measurable attributes. The remainder of this section introduces each component, its role within the layer, and its formal structure.

\paragraph{Topology Graph.} The topology graph is a live model of the entities in the managed environment and the structural relationships between them. Its role is to serve as the foundational skeleton over which all other reasoning is instantiated, providing agents with a continuously updated answer to "what exists in this environment and how is it connected?" Let $\mathcal{E} = \{e_1, \ldots, e_n\}$ be a finite set of entities, each representing a component of the environment such as a microservice, database, cache, workload, or infrastructure node. Relationships between entities are captured by the ternary relation
\[
\mathcal{R} \subseteq \mathcal{E} \times \mathcal{E} \times \{\textit{conn}, \textit{layer}, \textit{comp}\},
\]
where $(e_i, e_j, \textit{conn})$ denotes horizontal connectivity (e.g., one service calling another), $(e_i, e_j, \textit{layer})$ denotes a vertical layering relationship (e.g., a service running atop a workload, pod, or node), and $(e_i, e_j, \textit{comp})$ denotes compositional containment (e.g., a workload composed of pods). The topology graph is the tuple $\mathcal{G}_T = (\mathcal{E},\, \mathcal{R})$, maintained continuously as the environment changes.

\paragraph{Causal Knowledge Base.} The causal knowledge base is a global, environment-agnostic library encoding domain expertise about failure modes in managed cloud environments: what root causes can exist for each class of entity, what symptoms they produce, and how their effects propagate across relationship types. Its role is to separate durable domain knowledge, which applies broadly to cloud-native systems, from transient environment state, which differs per deployment, so that the layer can reason about any newly observed environment without requiring per-environment modeling. Let $\mathcal{T}$ be a set of entity types, $\mathcal{R}_c^*$ a universe of possible root causes, and $\mathcal{S}^*$ a universe of possible symptoms. The causal knowledge base is defined as
\[
\mathcal{K}_C = (\mathcal{T},\, \mathcal{R}_c^*,\, \mathcal{S}^*,\, \rho,\, \pi),
\]
where $\rho: \mathcal{T} \rightarrow 2^{\mathcal{R}_c^* \times \mathcal{S}^*}$ maps each entity type to the root-cause-to-symptom associations defined for it, and $\pi$ is a set of propagation rules describing how effects traverse the relation types in $\mathcal{R}$ (for example, how a symptom on a downstream service propagates over $\textit{conn}$ relationships to its callers) \cite{causelydocs}.

\paragraph{Causality Graph.} The causality graph is the environment-specific instantiation of the causal knowledge base over the topology graph and attribute dependency graph. Its role is to provide a concrete, probabilistic model of how root causes produce symptoms in this particular environment, supporting abductive inference: given a set of active symptoms, infer the most likely root causes. Let $\mathcal{R}_c \subseteq \mathcal{R}_c^*$ be the root causes instantiated for entities present in $\mathcal{E}$, and let $\mathcal{S} \subseteq \mathcal{S}^*$ be the symptoms instantiated for those entities. The causality graph is the directed acyclic graph
\[
\mathcal{G}_C = (V,\, E_C,\, P), \quad V = \mathcal{R}_c \cup \mathcal{S},
\]
where $E_C \subseteq V \times V$ are directed causal edges derived by applying the propagation rules $\pi$ of $\mathcal{K}_C$ to the topology relations in $\mathcal{R}$, and $P: E_C \rightarrow (0, 1]$ assigns to each edge $(r_i, s_j)$ the conditional probability $P(s_j \mid r_i)$ that root cause $r_i$ will produce symptom $s_j$.

\paragraph{Attribute Dependency Graph.} The attribute dependency graph captures functional relationships between measurable performance attributes across entities, such as latency, throughput, utilization, and queue length. Its role is to enable reasoning about how perturbations propagate across attributes and whether operational constraints remain satisfied, complementing the discrete cause-and-effect reasoning of the causality graph with continuous-valued analysis. Let $\mathcal{A} = \{a_1, \ldots, a_\ell\}$ be a set of attributes defined over entities in $\mathcal{E}$. The attribute dependency graph is the directed acyclic graph
\[
\mathcal{G}_A = (\mathcal{A},\, E_A,\, f),
\]
where $E_A \subseteq \mathcal{A} \times \mathcal{A}$ are directed dependency edges and $f$ labels each edge with a functional relationship, either predefined in the domain model or learned from observed behavior, such that a dependent attribute is expressed as a function of its parents.

By supplying $\mathcal{CI}$ directly to AI agents as structured context, a causal intelligence layer addresses the semantic-interpretation tax identified in Section 1.2: agents no longer need to infer topology, dependencies, or causal relationships from raw telemetry on every query. This should reduce token consumption, improve response latency, and increase diagnostic reliability. These are hypotheses we evaluate empirically in the remainder of this paper.

\subsection{Use Cases}

We organize the study around four use case categories that capture the principal reliability use cases found in practice: \textit{health assessment}, \textit{impact analysis}, \textit{root cause localization}, and \textit{remediation}. Each category corresponds to a distinct way in which an agent can consume the causal intelligence layer $\mathcal{CI}$, and each can be expressed as an operation over its structured state.

\paragraph{Health Assessment.} Health assessment is the task of analyzing the current state of a managed environment, or a designated scope within it, from its active telemetry. Its role is to answer the first operational question an agent or operator typically asks: ``what is happening right now?'' Formally, for a scope $\mathcal{E}' \subseteq \mathcal{E}$, let $\mathcal{O}^+(\mathcal{E}')$ denote the set of active observations drawn from the telemetry of entities in scope. Health assessment computes a structured interpretation
\[
\mathcal{H}(\mathcal{E}') \;=\; \bigl(\mathcal{S}^+(\mathcal{E}'),\; \mathcal{R}_c^+(\mathcal{E}')\bigr),
\]
where $\mathcal{S}^+(\mathcal{E}') \subseteq \mathcal{S}$ is the set of active symptoms inferred from $\mathcal{O}^+(\mathcal{E}')$, and $\mathcal{R}_c^+(\mathcal{E}') \subseteq \mathcal{R}_c$ is the set of root causes abductively supported by those symptoms within the scope. In this formulation, health assessment is not merely the detection of isolated anomalies, but the interpretation of environment state from telemetry into semantically meaningful operational conditions.

\paragraph{Incident Impact Analysis.} The incident impact analysis determines the scope of entities, services, and teams affected by an active symptom set or by a localized root cause. Its role is to answer questions of the form ``what is affected, and how broadly?'' Formally, given a root cause $r \in \mathcal{R}_c$, its directly expressed symptoms are
\[
\textit{effects}(r) \;=\; \{\, s \in \mathcal{S} : (r, s) \in E_C\,\},
\]
and the set of impacted entities is
\[
\mathcal{E}(r) \;=\; \{\, e \in \mathcal{E} : \exists\, s \in \textit{effects}(r) \text{ defined over } e\,\}.
\]
The broader blast radius is obtained by propagating these effects over the topology relations in $\mathcal{R}$ according to the propagation rules $\pi$, yielding the set of entities transitively affected by the failure. Organizational impact questions, such as whether multiple teams must be involved, reduce to projecting this impacted entity set onto ownership metadata.

\paragraph{Root Cause Localization.} Root cause localization is the task of identifying where in the environment the active failure is most likely originating. Its role is to answer questions such as ``what service is responsible?'' or ``is this our team's fault?'' Formally, given an active symptom set $\mathcal{S}^+$, the causal intelligence layer performs abductive inference over the causality graph $\mathcal{G}_C = (V, E_C, P)$ to select the most likely explanatory root cause
\[
r^* \;=\; \arg\max_{r \in \mathcal{R}_c} P(r \mid \mathcal{S}^+)
\;=\; \arg\max_{r \in \mathcal{R}_c} P(\mathcal{S}^+ \mid r)\, P(r),
\]
with
\[
P(\mathcal{S}^+ \mid r) \;=\; \prod_{s \in \mathcal{S}^+} P(s \mid r)
\]
under the conditional independence assumptions encoded by $\mathcal{G}_C$. In the benchmark setting, this operation localizes the failure to the implicated root-cause entity or service; the agent can then use conventional evidence sources such as logs, events, versions, and recent changes to investigate the concrete defect mechanism. Ownership attribution is a projection of this localization step: for a team owning $\mathcal{E}_{\text{team}} \subseteq \mathcal{E}$, responsibility holds when $\mathcal{E}(r^*) \cap \mathcal{E}_{\text{team}} \neq \emptyset$ \cite{causelydocs}.

\paragraph{Remediation.} Remediation is the task of deciding whether a proposed corrective action addresses the localized cause of failure or merely one of its downstream manifestations. Its role is to answer questions such as ``is this action safe?'' and ``does this resolve the actual problem?'' Formally, let an action $a$ target an entity $e_a \in \mathcal{E}$. Given a localized root cause $r^*$, the action is causally aligned when $e_a \in \mathcal{E}(r^*)$, meaning that the action is directed at the entity or component hosting the failure source. Otherwise, the action operates on a downstream affected entity and may suppress symptoms without resolving the underlying defect. In this sense, remediation over $\mathcal{CI}$ is not simply action recommendation, but action evaluation relative to the causal locus of the incident.

These four categories provide the conceptual frame for the benchmark query catalog. Framing the study in this way allows us to evaluate not only whether agents produce correct answers, but whether access to $\mathcal{CI}$ improves performance on the major classes of operational reasoning tasks that arise in continuous reliability work.

%==============================================================================
% SECTION 3: KEY FINDINGS
%==============================================================================

\section{Key Findings}
\label{sec:findings}

\begin{figure}[H]
\centering
\includegraphics[width=\linewidth]{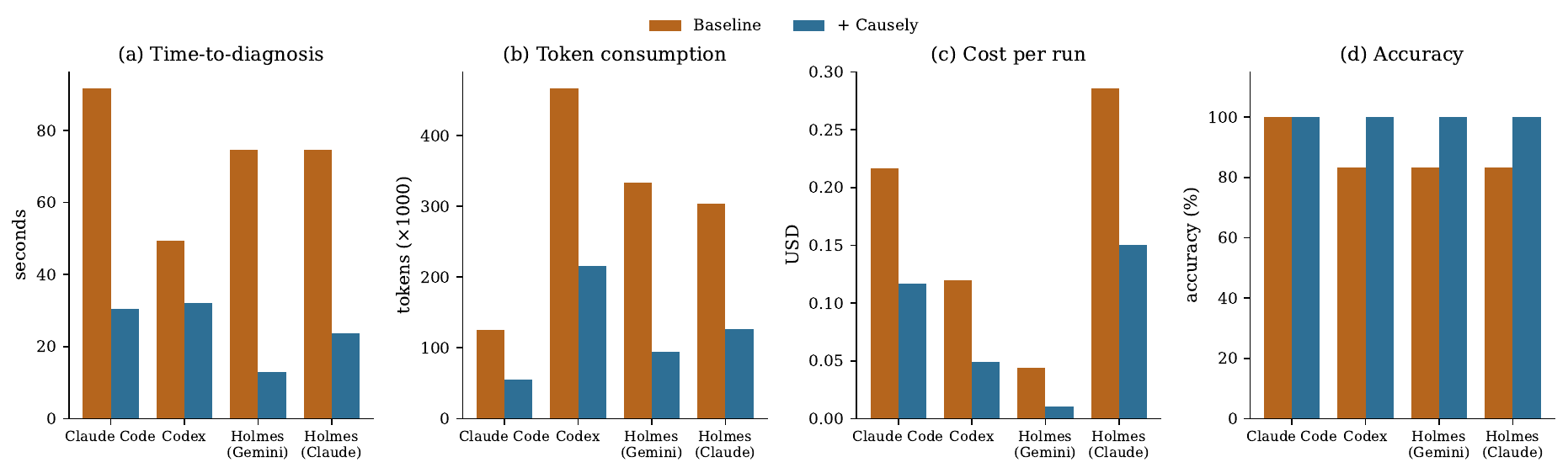}
\caption{Active-fault per-configuration summary: (a) time-to-diagnosis, (b) total tokens per run, (c) USD cost per run, and (d) accuracy. All four configurations improve on every dimension under causal grounding.}
\label{fig:sc2-overview}
\end{figure}

Across 72 runs, access to Causely's causal intelligence layer substantially improved every configuration across all evaluated metrics. On the active-fault scenario, mean time-to-diagnosis fell by 63\%, mean token consumption per run fell by 60\%, and the tool-call footprint of each investigation compressed by 4.8$\times$, driving a 57\% reduction in direct API cost per run. Accuracy rose from 75\% to 100\% on root cause diagnosis, the category for which baseline agents spent the most context and performed least reliably. On the healthy baseline, hallucinated-incident behavior was eliminated for HolmesGPT (Claude Sonnet) and halved for Codex. An inverted resource pattern also emerged: baseline agents consumed more time and tokens investigating the healthy cluster than the active fault, by ratios up to 7.2$\times$, because raw telemetry provided no signal that the correct answer was ``nothing is wrong.'' The remainder of this section explores these results along four axes: query time (\S\ref{sec:find-latency}), diagnostic accuracy (\S\ref{sec:find-accuracy}), token consumption and tool-call footprint (\S\ref{sec:find-tokens}), and cost (\S\ref{sec:find-cost}); limitations and anomalies are treated in \S\ref{sec:find-limits}. Figure~\ref{fig:sc2-overview} summarizes the four headline metrics on the active-fault scenario.

\subsection{Query Time Analysis}
\label{sec:find-latency}

Mean time-to-diagnosis under the active-fault scenario declined by \textbf{63.2\%} across the four configurations, with per-configuration reductions ranging from 34.8\% (Codex) to 82.8\% (HolmesGPT (Gemini Pro 3)). In one experiment, a HolmesGPT (Gemini Pro 3) run that took 281s without causal grounding completed in 13s with access to Causely.

\begin{table}[H]
\centering
\caption{Mean time-to-diagnosis under the active-fault scenario.}
\label{tab:speed}
\begin{tabular}{l|c|c|c}
\toprule
\textbf{Configuration} & \textbf{Base (s)} & \textbf{+Causely (s)} & \textbf{\% Change} \\
\midrule
Claude Code& 91.7 & 30.5 & $-66.7$\% \\
Codex& 49.3 & 32.2 & $-34.8$\% \\
HolmesGPT (Gemini Pro 3)& 74.5 & 12.8 & $-82.8$\% \\
HolmesGPT (Claude Sonnet)& 74.7 & 23.7 & $-68.3$\% \\
\midrule
\textbf{Productivity AI (mean)} & 70.5 & 31.3 & $-55.6$\% \\
\textbf{Ops AI (mean)}          & 74.6 & 18.2 & $-75.5$\% \\
\bottomrule
\end{tabular}
\end{table}

The latency gain arose from replacing open-ended interpretation of environment state with targeted causal lookup (\S\ref{sec:find-toolcalls}). Baseline agents constructed a view of the environment through repeated CLI or generic MCP calls. Agents with access to Causely instead issued a small number of queries against pre-computed causal structure, such as \texttt{get\_root\_causes}, \texttt{get\_environment\_health}, and \texttt{get\_symptoms}, and received already-resolved causal state.

\begin{figure}[H]
\centering
\includegraphics[width=\linewidth]{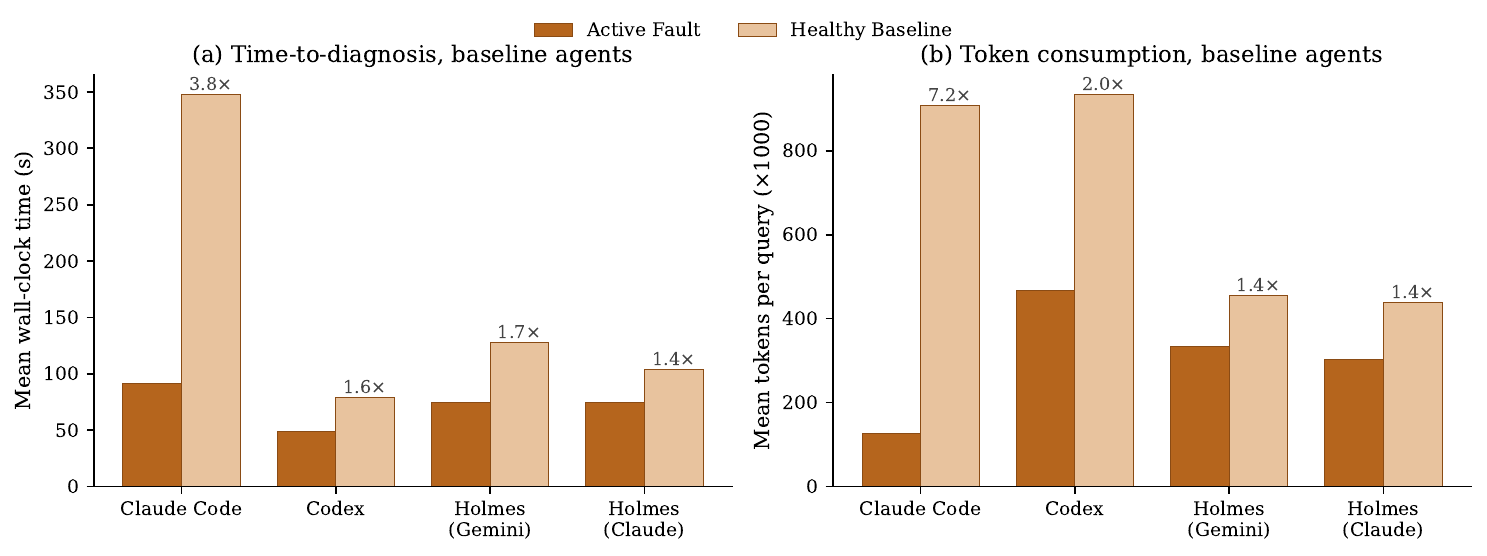}
\caption{Baseline agents expend more time and tokens on the healthy baseline than on the active fault. Annotated ratios compare healthy-baseline to active-fault consumption per configuration. Absence of fault, rather than the fault itself, is the expensive case for agents operating directly on raw telemetry.}
\label{fig:paradox}
\end{figure}

The same mechanism also explains a second pattern visible when baseline latency is broken out by scenario (Figure~\ref{fig:paradox}). \emph{Every} baseline configuration spent more time, and substantially more tokens, investigating the healthy baseline than the active fault. Claude Code was the most extreme case, averaging 348s and 909K tokens per healthy-baseline query, corresponding to 3.8$\times$ and 7.2$\times$ inflation, respectively, relative to its active-fault runs. Raw telemetry provided no principled stopping condition for a reasoning loop asked whether anything was wrong. In the absence of a causal ``no active root cause'' signal, agents continued searching until some pattern could be interpreted as an incident. Causal grounding reversed this pattern because absence of fault became a first-class answer rather than a failure to find one.

The latency reduction comes from reducing the number of sequential reasoning and tool-use steps required to answer the query. In the baseline setting, agents build an answer incrementally: each observation informs the next command or tool call, so total query time accumulates across repeated round trips and intermediate reasoning steps. With access to the causal intelligence layer, much of that intermediate state is already assembled. Topology, active symptoms, and likely root causes can be retrieved directly through a small number of causal queries rather than reconstructed through repeated CLI probes. The reduction is therefore largest for the configurations whose baseline behavior depended most heavily on iterative exploration.

\subsection{Diagnostic Accuracy Analysis}
\label{sec:find-accuracy}

Under the active-fault scenario, causal grounding eliminated missed diagnoses in every configuration that exhibited them at baseline (Table~\ref{tab:accuracy}). Codex, HolmesGPT (Gemini Pro 3), and HolmesGPT (Claude Sonnet) each scored 83\% at baseline; all four configurations reached 100\% accuracy under causal access.

\begin{table}[H]
\centering
\caption{Active-fault diagnostic accuracy.}
\label{tab:accuracy}
\begin{tabular}{l|c|c}
\toprule
\textbf{Configuration} & \textbf{Base} & \textbf{+Causely} \\
\midrule
Claude Code& 100\% & 100\% \\
Codex&  83\% & 100\% \\
HolmesGPT (Gemini Pro 3)&  83\% & 100\% \\
HolmesGPT (Claude Sonnet)&  83\% & 100\% \\
\bottomrule
\end{tabular}
\end{table}

The clearest accuracy effect appeared on the healthy baseline. In this setting, the relevant failure mode was a false positive rather than a missed diagnosis. The agent hallucinated an incident even though no active fault was present. HolmesGPT (Claude Sonnet) and Codex both produced false positives at a 67\% rate in the baseline condition (Table~\ref{tab:fp}). Access to Causely eliminated this behavior for HolmesGPT (Claude Sonnet) and reduced the false-positive rate for Codex to 33\%. Claude Code and HolmesGPT (Gemini Pro 3) produced no false positives under either condition.

\begin{table}[H]
\centering
\caption{Hallucinated-incident rate under the healthy baseline.}
\label{tab:fp}
\begin{tabular}{l|c|c}
\toprule
\textbf{Configuration} & \textbf{Base FP Rate} & \textbf{+Causely FP Rate} \\
\midrule
Claude Code&  0\% &  0\% \\
Codex& 67\% & 33\% \\
HolmesGPT (Gemini Pro 3)&  0\% &  0\% \\
HolmesGPT (Claude Sonnet)& 67\% &  0\% \\
\bottomrule
\end{tabular}
\end{table}

The baseline hallucinations followed the same basic pattern: the agent constructed an incident narrative without a grounded causal explanation. Three representative failures from the evaluator-scored run traces illustrate the pattern:
\begin{itemize}\itemsep1pt
  \item \textit{A causal chain built from stale telemetry.} Asked for the root cause of a non-existent incident, Codex traced \texttt{ECONNREFUSED} errors in the checkout logs to a product-catalog pod restart that had occurred \emph{74 minutes earlier}, turning residual historical signals into a plausible but incorrect incident narrative.
  \item \textit{An incident inferred from a tooling artifact.} A TCP connectivity probe issued by HolmesGPT (Claude Sonnet) failed because \texttt{kubectl} was unavailable in the agent's temporary working directory. The agent interpreted this as ``checkout not responding on port 8080'' and then built a multi-service blast-radius narrative on top of that false premise.
  \item \textit{Misattribution of unrelated errors.} OTel Collector TLS validation failures against \texttt{kubelet} were genuine in the telemetry but causally unrelated to the checkout path. HolmesGPT (Claude Sonnet) nonetheless attributed the alerted errors to them.
\end{itemize}

\begin{figure}[H]
\centering
\includegraphics[width=0.82\linewidth]{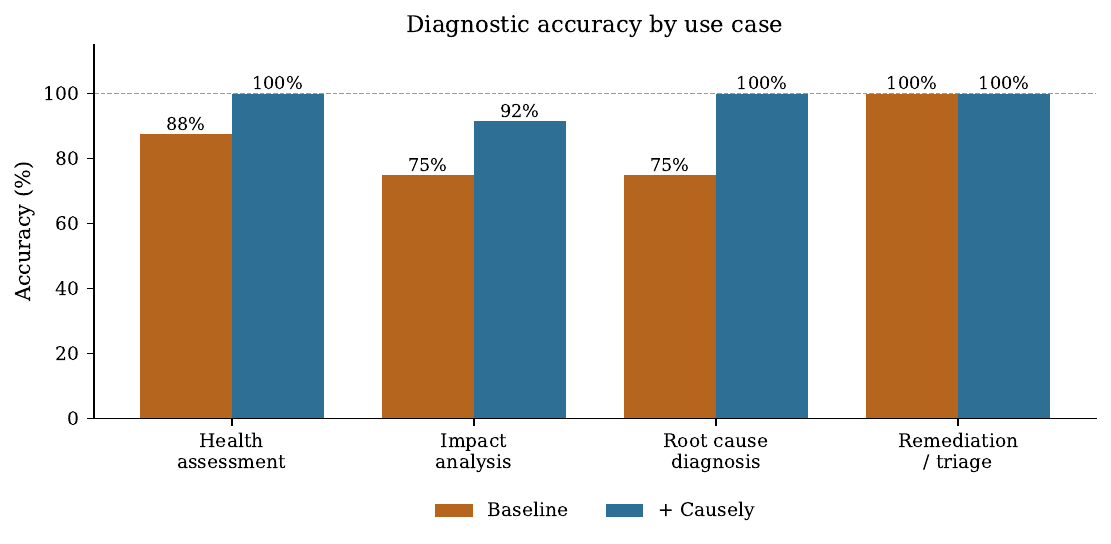}
\caption{Diagnostic accuracy by query category, pooled across configurations and both scenarios. The largest absolute and relative gains are on root-cause diagnosis, the category with the lowest baseline accuracy.}
\label{fig:acc-cat}
\end{figure}

Diagnostic accuracy differed substantially by use case (Figure~\ref{fig:acc-cat}, Table~\ref{tab:acc-cat}). Baseline diagnostic accuracy was 100\% on remediation queries and 87.5\% on health assessment. Impact analysis and root cause diagnosis were lower, at \textbf{75\%} in both categories.  These were the two categories that required combining evidence across multiple entities. Root cause diagnosis also consumed the most resources at baseline, with 694K mean tokens and 148s mean time-to-diagnosis. In other words, baseline agents spent the most context on the category in which they were least reliable. Under causal grounding, diagnostic accuracy reached 100\% on health assessment, root cause diagnosis, and remediation, and 91.7\% on impact analysis.

\begin{table}[H]
  \centering
  \small
  \caption{Per-use-case diagnostic accuracy, mean time-to-correct-diagnosis (TTCD, in seconds), and mean token consumption, pooled across configurations and scenarios, with and without access to Causely. Relative percentage differences in the +Causely columns are computed against the baseline condition.}
  \label{tab:acc-cat}
  \renewcommand{\arraystretch}{1.1}
  \begin{tabular}{l|cc|cc|cc}
  \toprule
  \textbf{Use case} & \multicolumn{2}{c|}{\textbf{Accuracy}} & \multicolumn{2}{c|}{\textbf{Mean TTCD (s)}} & \multicolumn{2}{c}{\textbf{Mean tokens}} \\
    & Base & +Causely & Base & +Causely & Base & +Causely \\
  \midrule
  Health assessment    & 87.5\%  & 100.0\% (+14.3\%) &  44 &  19 (-56.8\%) & 151K & 100K (-33.8\%) \\
  Impact analysis      & 75.0\%  &  91.7\% (+22.3\%) & 116 &  31 (-73.3\%) & 427K & 181K (-57.6\%) \\
  Root cause diagnosis & 75.0\%  & 100.0\% (+33.3\%) & 148 &  57 (-61.5\%) & 694K & 351K (-49.4\%) \\
  Remediation          & 100.0\% & 100.0\% (+0.0\%)  &  50 &  29 (-42.0\%) & 233K & 176K (-24.5\%) \\
  \bottomrule
  \end{tabular}
\end{table}

Causely improves diagnostic accuracy because it removes the need for Ops AI agents to reconstruct causal state from raw telemetry and instead provides direct access to pre-computed causal structure. Impact analysis and root cause localization required combining evidence across multiple entities and distinguishing causal drivers from downstream effects. Root cause localization was the hardest case because the Ops AI agent had to work backward from an observed symptom pattern to the responsible service or entity. With only logs and metrics as inputs, there was often no principled basis for preferring one plausible explanation over another. The causal graph provided that missing structure by identifying the service or entity most likely responsible for the observed symptoms. The same point mattered on the healthy baseline. There, the relevant failure mode was the false positive: the Ops AI agent hallucinated an incident because raw telemetry provided no grounded representation of the absence of fault. A ``no active root cause'' result changed that by giving the agent a concrete basis for concluding that no incident was present.

\subsection{Token Consumption and Tool Call Analysis}
\label{sec:find-tokens}
\label{sec:find-toolcalls}

Mean token consumption per run under the active-fault scenario declined by \textbf{59.9\%}, and mean tool-call count declined by \textbf{78.4\%}, corresponding to a mean investigation-footprint compression of 4.8$\times$. Both reductions appeared in every configuration (Tables~\ref{tab:tokens} and~\ref{tab:toolcalls}). Grouped by agent type, productivity agents reduced from a mean of 296K to 136K tokens per run ($-54.1$\%), and Ops agents from 319K to 110K ($-65.5$\%). Tool-call compression was also larger for Ops agents than for productivity agents (5.5$\times$ versus 4.2$\times$). This difference reflects the structure of the baseline interfaces: the Ops baseline already operated through a typed interface, but it still required a larger number of narrow queries than the denser queries exposed by the causal layer.

\begin{table}[H]
\centering
\small
\caption{Token consumption under the active-fault scenario.}
\label{tab:tokens}
\setlength{\tabcolsep}{4pt}
\renewcommand{\arraystretch}{1.1}
\begin{tabular}{l|c|c|c|c|c}
\toprule
\textbf{Configuration} & \textbf{Avg Base} & \textbf{Avg +C} & \textbf{Max Base} & \textbf{Max +C} & \textbf{\% Change} \\
\midrule
Claude Code& 126K & 56K  & 278K & 58K  & $-55.7$\% \\
Codex& 467K & 216K & 615K & 456K & $-53.7$\% \\
HolmesGPT (Gemini Pro 3)& 334K & 94K  & 813K & 111K & $-71.7$\% \\
HolmesGPT (Claude Sonnet)& 304K & 126K & 416K & 158K & $-58.4$\% \\
\midrule
\textbf{Productivity AI (mean)} & 296K & 136K & -- & -- & $-54.1$\% \\
\textbf{Ops AI (mean)}          & 319K & 110K & -- & -- & $-65.5$\% \\
\bottomrule
\end{tabular}
\end{table}

Worst-case token consumption compressed more sharply than the mean: the largest single baseline run (HolmesGPT (Gemini Pro 3), 813K tokens) resolved in 111K under causal grounding, an 86\% reduction at the envelope. The upper envelope is the relevant figure for context-window pressure and provider rate limits, and it is the envelope along which causal grounding delivers its largest absolute gains. Per-configuration tool-call compression ranges from 3.4$\times$ (Claude Code) to 6.1$\times$ (HolmesGPT (Claude Sonnet)).

\begin{table}[H]
\centering
\small
\caption{Tool invocations per investigation under the active-fault scenario.}
\label{tab:toolcalls}
\renewcommand{\arraystretch}{1.1}
\begin{tabular}{l|c|c|c|c|c}
\toprule
\textbf{Configuration} & \textbf{Avg Base} & \textbf{Avg +C} & \textbf{Max Base} & \textbf{Max +C} & \textbf{\% Change} \\
\midrule
Claude Code& 13.7 & 4.0 & 29 &  4 & $-70.7$\% \\
Codex& 22.0 & 4.5 & 33 &  8 & $-79.5$\% \\
HolmesGPT (Gemini Pro 3)& 16.3 & 3.3 & 28 &  4 & $-79.6$\% \\
HolmesGPT (Claude Sonnet)& 23.3 & 3.8 & 30 &  5 & $-83.6$\% \\
\midrule
\textbf{Productivity AI (mean)} & 17.8 & 4.3 & -- & -- & $-76.2$\% \\
\textbf{Ops AI (mean)}          & 19.8 & 3.6 & -- & -- & $-82.2$\% \\
\bottomrule
\end{tabular}
\end{table}

\begin{figure}[H]
\centering
\begin{minipage}{0.46\linewidth}
\centering
\includegraphics[width=\linewidth]{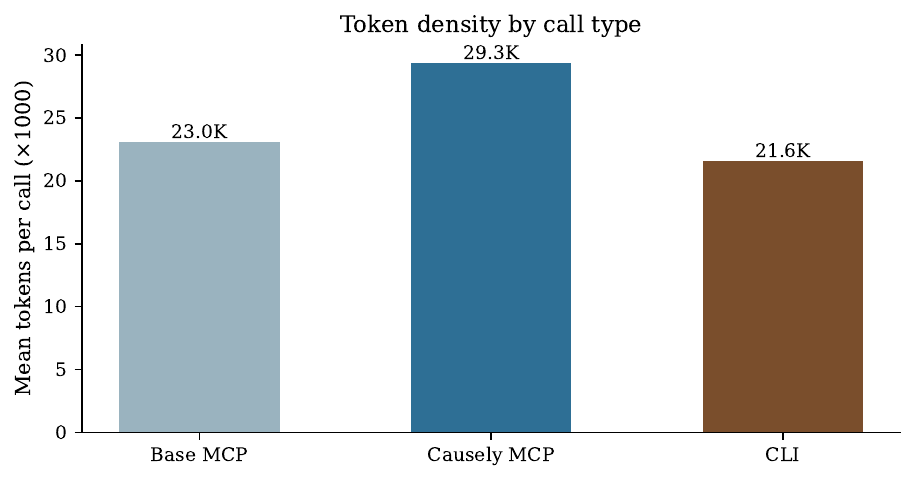}
\end{minipage}\hfill
\begin{minipage}{0.50\linewidth}
\centering
\includegraphics[width=\linewidth]{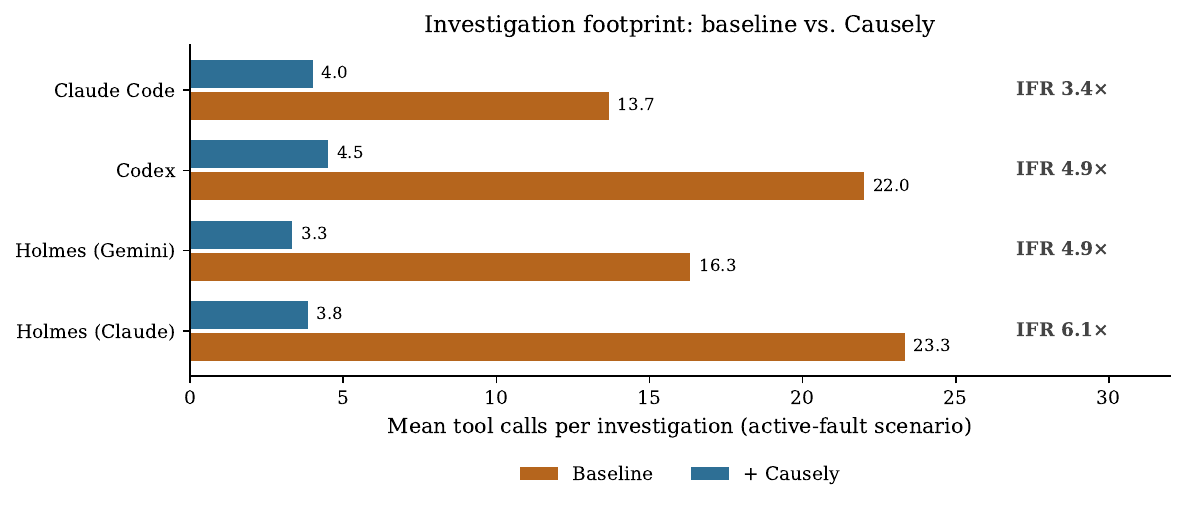}
\end{minipage}
\caption{Left: mean tokens per call by call type; causal-layer calls are denser than generic MCP calls and smaller than CLI invocations. Right: Investigation Footprint Ratio (IFR), the compression in tool-call count between baseline and treatment conditions under the active-fault scenario.}
\label{fig:tpc-ifr}
\end{figure}

This pattern is explained by the token density of individual calls (Figure~\ref{fig:tpc-ifr}, left). Baseline CLI invocations consumed 46.9K tokens on average, driven by large contextual payloads such as pod listings, log pages, and trace snippets. Generic MCP calls consumed 15.6K tokens on average. Causely MCP calls consumed 31.5K tokens on average: roughly twice a generic MCP call and slightly more than half a CLI call. A Causely MCP call was therefore denser, not cheaper per invocation. Each call returned a pre-assembled slice of causal state, such as topology, active root causes, blast radius, or attribute dependencies, that a baseline agent would otherwise have reconstructed through several smaller calls. The overall efficiency gain came from reducing the number of calls required per investigation. Figure~\ref{fig:tpc-ifr} (right) quantifies this reduction through the Investigation Footprint Ratio.

The same pattern also appeared in failed tool invocations. Productivity agents, which relied on free-form shell access, averaged between 1.5 and 3.5 failed tool calls per baseline run, driven by mistyped commands, missing binaries, and incorrect paths. Under causal grounding, failed calls fell to zero on the active-fault scenario for both productivity configurations. Ops agents, which already operated through a typed MCP interface, exhibited essentially zero failed calls in either condition. The causal interface therefore improved not only performance but also robustness. Calls that would otherwise have failed as malformed shell commands were replaced by well-formed causal queries.

The reduction in tokens and tool calls came from shifting environment interpretation work out of the Ops AI agent's inference loop. In the baseline setting, the agent had to convert raw telemetry into a usable representation of environment state at run time. That process included enumerating pods, correlating log lines with metric deviations, reconstructing service relationships from trace metadata, and inferring a propagation path. Each step consumed tokens because the intermediate state had to be rebuilt inside the agent's context window. The causal intelligence layer performed this work ahead of time and exposed the result as typed objects such as entities, symptoms, root causes, and propagation edges. A single causal query could therefore replace the accumulated output of many smaller reconstructive calls. This is why total token use fell even though an individual Causely call was denser than a generic MCP call. The largest compression appeared in the configurations whose baseline workflow depended most heavily on repeated reconstruction of environment state. The same effect appeared for both productivity and Ops AI agents because both had to pay that reconstruction cost when operating directly on raw telemetry.

\subsection{Token Economics and Cost Analysis}
\label{sec:find-cost}

The cost results follow directly from the token findings in Section~\ref{sec:find-tokens}. We evaluated per-run API cost from the token usage recorded by the benchmark harness, including input, cached-input, and output token charges; engineer labor, infrastructure, and fixed platform licensing are excluded throughout. The benchmark ran on lower-cost model tiers (Claude Sonnet, Gemini 3.1 Flash-Lite, and GPT-5.4 mini) to keep experimental overhead bounded. Under those benchmark-tier prices, mean per-run API cost on the active-fault scenario declined by \textbf{56.9\%} across configurations (Table~\ref{tab:cost}).

\begin{table}[H]
\centering
\small
\caption{Direct API cost per active-fault run under the benchmark model tiers. Cost/run denotes mean API cost per investigation run.}
\label{tab:cost}
\renewcommand{\arraystretch}{1.1}
\begin{tabular}{l|l|ccc}
\toprule
\textbf{Configuration} & \textbf{Benchmark model} & \textbf{Base} & \textbf{+Causely} & \textbf{\% change} \\
\midrule
Claude Code & Claude Sonnet & \$0.216 & \$0.117 & $-46.0\%$ \\
Codex & GPT-5.4 mini & \$0.119 & \$0.049 & $-58.6\%$ \\
HolmesGPT (Gemini Pro 3) & Gemini 3.1 Flash-Lite & \$0.044 & \$0.011 & $-75.7\%$ \\
HolmesGPT (Claude Sonnet) & Claude Sonnet & \$0.286 & \$0.150 & $-47.4\%$ \\
\midrule
\textbf{Productivity AI (mean)} & -- & \$0.168 & \$0.083 & $-50.5\%$ \\
\textbf{Ops AI (mean)} & -- & \$0.165 & \$0.081 & $-51.1\%$ \\
\bottomrule
\end{tabular}
\end{table}

To characterize the cost differential at higher-capability model tiers, we repriced the same active-fault token traces at current published standard API rates for Claude Opus 4.7, GPT-5.5, and Gemini 3.1 Pro. For GPT-5.5, we used standard API pricing; reasoning effort does not have a separate per-token price. For Gemini 3.1 Pro, we applied the documented higher price tier for prompts exceeding 200K tokens.

\begin{table}[H]
\centering
\small
\caption{Premium-tier sensitivity analysis. The same active-fault token traces are repriced using current standard API rates for higher-cost model tiers.}
\label{tab:cost-premium}
\renewcommand{\arraystretch}{1.1}
\begin{tabular}{l|l|ccc}
\toprule
\textbf{Configuration} & \textbf{Premium model} & \textbf{Base} & \textbf{+Causely} & \textbf{\% change} \\
\midrule
Claude Code & Claude Opus 4.7 & \$0.360 & \$0.195 & $-46.0\%$ \\
Codex & GPT-5.5 & \$1.496 & \$0.678 & $-54.7\%$ \\
HolmesGPT (Gemini Pro 3) & Gemini 3.1 Pro & \$1.348 & \$0.199 & $-85.2\%$ \\
HolmesGPT (Claude Sonnet) & Claude Opus 4.7 & \$0.476 & \$0.251 & $-47.4\%$ \\
\midrule
\textbf{Productivity AI (mean)} & -- & \$0.928 & \$0.436 & $-53.0\%$ \\
\textbf{Ops AI (mean)} & -- & \$0.912 & \$0.225 & $-75.3\%$ \\
\bottomrule
\end{tabular}
\end{table}

Across both tables, the relative cost reduction is consistent: causal grounding cuts per-run API cost by 47--76\% at benchmark-tier prices and 46--85\% at premium-tier prices. What changes between the two is the absolute magnitude. At benchmark-tier prices, per-run savings range from \$0.03 to \$0.14. At premium-tier prices, the same token reduction yields \$0.17--\$1.15 per run. All figures are direct API cost and exclude engineer labor. They also understate the full baseline burden: as noted in Section~\ref{sec:find-latency}, baseline agents consumed up to 7.2$\times$ more tokens on a healthy cluster than on an active fault, so the active-fault cost figures here are a lower bound on what baseline agents actually spend in production.

%==============================================================================
% SECTION 3: METHODOLOGY
%==============================================================================

\section{Methodology}
\label{sec:methodology}

We evaluate the research question posed in Section~1.4 via a fully crossed factorial benchmark over two independent variables: \textit{agent configuration} (four levels, spanning the productivity and Ops archetypes) and \textit{causal access} (two levels, baseline versus baseline plus the Causely MCP server that exposes the causal intelligence layer $\mathcal{CI}$ defined in Section~2). Each of the resulting eight cells is evaluated under two operational scenarios: an active-fault scenario, in which a code-level defect is injected into the payment service of the target application, and a healthy-baseline scenario, in which the same application runs without faults. The active-fault scenario administers six queries per cell and the healthy-baseline scenario administers three, yielding 72 experimental runs in total. Within a configuration, prompts, models, and permissions are held constant across conditions, so that any observed difference is attributable to the presence or absence of access to $\mathcal{CI}$. The detailed enumeration of configurations, queries, and the correctness rubric appears in Appendix~\ref{app:setup}.

Each sub-question of RQ is operationalized by one or more quantitative measurements recorded on every run. Table~\ref{tab:metrics} defines each metric, the sub-question it addresses, and the measurement procedure.

\begin{table}[H]
\centering
\caption{Metrics collected per experimental run, the research sub-question each addresses, and the measurement procedure. All metrics are recorded under both the baseline and treatment conditions, enabling paired comparison per configuration.}
\label{tab:metrics}
\small
\renewcommand{\arraystretch}{1.15}
\begin{tabular}{p{2.5cm} p{0.9cm} p{4.3cm} p{5.2cm}}
\toprule
\textbf{Metric} & \textbf{Addr.} & \textbf{Definition} & \textbf{Measurement} \\
\midrule
Wall-clock time (s)          & RQ.1  & End-to-end latency from query submission to final agent response. & Timestamped by the orchestration harness; reported in seconds per run. \\
Correctness                  & RQ.2  & Binary indicator of whether the response satisfies the scenario-specific ground-truth rubric. & Scored against the rubric in Appendix~\ref{app:setup}; aggregated as accuracy under the active-fault scenario and false positive rate under the healthy baseline. \\
Token consumption            & RQ.3  & Total tokens billed against the underlying model per run (input, output, cached). & Read directly from the model provider's per-response usage fields. \\
\midrule
Tool invocations             & ---   & Count of tool calls issued per run, stratified by tool category. & Counted from the recorded execution trace; supporting metric characterizing \emph{how} tokens are consumed. \\
Query cost (USD)             & ---   & Estimated API cost for a single run, derived from token consumption. & $c = p_\text{in} \cdot n_\text{in} + p_\text{out} \cdot n_\text{out}$ at published per-token pricing; derived economic projection of RQ.3. \\
\bottomrule
\end{tabular}
\end{table}

The three metrics above the rule directly address RQ.1--RQ.3. The two below are derived supporting metrics: tool invocations characterize the interaction pattern through which tokens are consumed, and query cost translates token consumption into an operational economic figure. They inform the findings but are not themselves objects of inquiry.

%==============================================================================
% SECTION 5: EXPERIMENT SETUP
%==============================================================================

\section{Experiment Setup}
\label{sec:setup}

\subsection{Environment and Application}

All experiments were conducted against the OpenTelemetry Astronomy Shop \cite{otel-demo}, a CNCF-maintained polyglot microservices reference application. The deployment comprised 24 microservices exercising six distinct infrastructure categories (Kubernetes workloads, PostgreSQL, Valkey/Redis, Kafka, OpenSearch, and a feature-flag service), providing a realistic cross-section of the entity types over which the causal intelligence layer operates. The application was deployed on a local \texttt{kind} cluster with native OpenTelemetry instrumentation producing traces, metrics, and logs. In treatment runs, the Causely mediator was deployed alongside the application and exposed to the agent via the Causely MCP server; baseline runs used the same cluster and observability stack without the Causely MCP server, holding the underlying raw telemetry constant across conditions.

\subsection{Summary of Experimental Protocol}

Each of the four agent configurations (two productivity AI agents, two Ops AI agents) was executed under two conditions (baseline and baseline plus Causely) against two scenarios: an active-fault scenario, in which a code-level defect in the payment service causes every transaction to be rejected and propagates degradation into three downstream services, and a healthy-baseline scenario, in which the same application runs without faults. The active-fault scenario administers six queries per cell and the healthy-baseline scenario administers three, yielding 72 experimental runs. For each run the execution trace, token usage, tool invocations, and final response were persisted and scored against a pre-registered correctness rubric. Full details of the agent configurations, the injected fault, the correctness rubric, and the data-collection schema are given in Appendix~\ref{app:setup}.

\section{Conclusion}

Across our experiments, we find that access to Causely improved agent performance on every measured dimension: mean time-to-diagnosis declined by 63.2\%, token consumption by 59.9\%, root-cause diagnosis accuracy rose from 75\% to 100\%, and direct API cost per run declined by 56.9\% at benchmark-tier prices. The improvement held across all four configurations and both operational scenarios. The same mechanism explains both results. Without causal grounding, agents derive environment state from raw telemetry at query time, enumerating services, correlating logs across service boundaries, and reconstructing a propagation path before any diagnostic reasoning can begin. Investigation cost accumulates in proportion to how much an agent must probe before it can answer. The healthy-baseline results make this explicit: baseline agents spent between 1.4$\times$ and 7.2$\times$ more time and tokens on a healthy cluster than on an active fault, because raw telemetry provides no stopping point for a reasoning loop asked whether anything is wrong. Without a ``no active root cause'' signal, agents continued searching until some pattern could be framed as an incident. Causal grounding eliminated this in both cases by supplying a typed answer, either a root cause or an explicit negative result, and removing the reconstruction work required to reach it.

\subsection{Discussion}

The benchmark measured the reconstruction tax on a 24-service environment under a single injected fault. These are favorable conditions for baseline agents: a small topology, an isolated fault, and a service count that limits how much cross-service correlation any investigation requires. In production, each of these factors scales up. To examine what the cost structure looks like as environments grow, we project from the benchmark Ops AI mean token counts across three representative environment profiles. We assume baseline token consumption scales linearly with service count, which is a conservative lower bound since multi-service incidents in larger environments require proportionally more cross-service log and trace correlation. We assume Causely token consumption scales with active issue count rather than topology size, since pre-computed causal state is returned in compact form regardless of how many services the environment contains.

\begin{table}[H]
\centering
\small
\caption{Projected token consumption and API cost per investigation. The Benchmark column is measured (Ops AI mean); Mid-tier and Enterprise columns are projected under the stated scaling assumptions using premium-tier rates. All cost figures exclude engineer labor.}
\label{tab:scale}
\renewcommand{\arraystretch}{1.1}
\begin{tabular}{l|c|c|c}
\toprule
 & \textbf{Benchmark} & \textbf{Mid-tier} & \textbf{Enterprise} \\
 & \textbf{24 services} & \textbf{$\sim$200 services} & \textbf{$\sim$1{,}000 services} \\
\midrule
Baseline tokens             & 319K   & $\sim$2.5M      & $\sim$12.8M \\
+Causely tokens             & 110K   & $\sim$220K      & $\sim$330K  \\
Baseline cost/run           & \$0.91 & $\sim$\$7.30    & $\sim$\$36  \\
+Causely cost/run           & \$0.23 & $\sim$\$0.45    & $\sim$\$0.68 \\
Cost reduction              & 75\%   & $\sim$94\%      & $\sim$98\%  \\
\midrule
Baseline, 500 runs/mo       & \$455  & $\sim$\$3{,}650 & $\sim$\$18{,}000 \\
+Causely, 500 runs/mo       & \$115  & $\sim$\$225     & $\sim$\$340 \\
\bottomrule
\end{tabular}
\end{table}

Baseline agents operating on raw telemetry are inefficient reasoners. Our benchmark records an average of 18 tool calls and 319K tokens per investigation, spent reconstructing environment state before any diagnostic reasoning begins. This reconstruction work is not diagnostic effort: it is overhead paid on every query to establish what the environment looks like before the reliability question can be addressed. It scales with topology complexity, not with the difficulty of the question being asked.

This cost is currently invisible to enterprises. Flat-rate seat subscriptions absorb token consumption at the provider layer, so an agent consuming 319K tokens per investigation looks identical on an invoice to one consuming 110K. That pricing model is not sustainable. OpenAI projects operating losses exceeding \$74 billion by 2028, driven primarily by compute spending. Anthropic has already moved enterprise customers from seat-based to per-token billing, a change reported to potentially triple costs for heavy consumers. The subsidy that makes the reconstruction tax invisible is ending.

When it ends, organizations that have built reliability workflows on token-inefficient agents carry a specific and quantifiable liability. Table~\ref{tab:scale} projects this under conservative linear scaling: multi-service incidents in real environments involve cascading failures where correlation work grows faster than linearly with topology size, so the actual baseline cost is likely higher than the table shows. Even at these conservative estimates, baseline cost reaches \$7.30 per investigation at 200 services, or approximately \$3,650 per month at 500 investigations. At 1,000 services, the baseline token estimate of 12.8M per investigation exceeds the context window of current frontier models, which cap at approximately 1 million tokens, by more than an order of magnitude. Beyond that threshold the failure is not financial. An agent that cannot fit its investigation into context produces a truncated result or fails outright. An organization that has committed substantial reliability infrastructure to raw-telemetry agents is not just paying more at scale. It is building operational dependency on agents that cannot complete investigations in the environments the business is growing into.

When the pricing model changes, two outcomes follow, and both can occur simultaneously. First, per-investigation costs escalate rapidly. An organization running 500 investigations per month at enterprise scale faces approximately \$18,000 per month in baseline model spend at premium-tier rates, a cost that does not appear on any invoice today under flat-rate subscriptions. Second, providers seeking to control compute costs may impose per-customer token limits. Under token limits, baseline agents face a direct choice between fewer investigations or incomplete ones: an investigation that cannot fit within a token budget is either truncated or does not run. Both outcomes degrade the reliability workflow. An organization that has replaced a substantial part of its reliability operations with AI agents cannot absorb a forced reduction in investigation capacity without operational impact. Causal grounding avoids both outcomes. By keeping per-investigation token consumption bounded regardless of environment size, it decouples the reliability workflow from the pricing risk that token-inefficient agents carry.

\subsection{Limitations and Future Work}
\label{sec:find-limits}

Several limitations qualify the generality of these findings. The active-fault scenario evaluates a single class of defect, an application-layer code bug, and does not directly speak to infrastructure faults, misconfigurations, or resource-exhaustion failures, each of which may exercise the causal intelligence layer differently. The deployment environment is a local \texttt{kind} cluster rather than a production-scale system, so absolute latency and token figures should be interpreted as relative comparisons within a controlled environment rather than as projections onto customer-scale workloads. Sample sizes per condition are modest (six queries for the fault scenario and three for the healthy baseline), and we do not report confidence intervals or formal significance tests; the large and consistent effect sizes observed across configurations mitigate but do not eliminate this concern. Prompt wording was fixed per query to isolate the treatment effect, but the results may be sensitive to phrasing, and no prompt-perturbation study was conducted. Finally, the HolmesGPT (Gemini Pro 3)configuration deviated from the otherwise consistent pattern of improvement on several metrics; the underlying cause is unresolved and is itself a subject for further investigation.

\appendix

\section*{Appendix}
\addcontentsline{toc}{section}{Appendix}

\section{Query Catalog}

\subsection{Healthy-Baseline Scenario}

\begin{enumerate}
\item What's the current health of the otel-demo namespace? Are there any active incidents or issues?
\item I'm seeing CheckoutServiceHighRequestErrors in the otel-demo namespace. What services are impacted and how widespread is it?
\item I'm seeing CheckoutServiceHighRequestErrors in the otel-demo namespace. What's the root cause?
\end{enumerate}

\subsection{Active-Fault Scenario}

\begin{enumerate}
\item What's the current health of the otel-demo namespace? Are there any active incidents or issues?
\item I'm seeing CheckoutServiceHighRequestErrors in the otel-demo namespace. What services are impacted and how widespread is it?
\item I'm seeing CheckoutServiceHighRequestErrors in the otel-demo namespace. What's the root cause?
\item My team owns the payment service. We got paged about checkout errors — is this our fault?
\item Is it safe to restart the payment pods, or is there a deeper issue we need to fix first?
\item Checkout is down. Is this a single team issue or do multiple teams need to be involved?
\end{enumerate}

\section{Experimental Setup Details}
\label{app:setup}

This appendix expands the summary in Section~\ref{sec:setup} with the full enumeration of configurations, the fault-injection procedure, the correctness rubric, and the data-collection schema.

\subsection{Factorial Structure}

\begin{table}[H]
\centering
\caption{Factorial structure of the benchmark. Each cell is evaluated under both the active-fault scenario (six queries) and the healthy-baseline scenario (three queries), yielding 72 experimental runs in total (4 configurations $\times$ 2 conditions $\times$ 9 queries).}
\label{tab:factorial}
\small
\begin{tabular}{lll}
\toprule
\textbf{Archetype} & \textbf{Configuration} & \textbf{Conditions} \\
\midrule
Productivity AI & Claude Code  & Baseline, Baseline + Causely \\
Productivity AI & Codex  & Baseline, Baseline + Causely \\
Ops AI          & HolmesGPT (Gemini Pro 3)  & Baseline, Baseline + Causely \\
Ops AI          & HolmesGPT (Claude Sonnet)  & Baseline, Baseline + Causely \\
\bottomrule
\end{tabular}
\end{table}

\subsection{Agent Configurations}

The four agent configurations vary agent framework and underlying model independently across the two archetypes. Productivity AI agents are deployed against their native toolset (shell access and \texttt{kubectl}), approximating how a general-purpose developer agent would be deployed in practice when asked to diagnose a reliability issue. Ops AI agents rely on a curated set of MCP servers for observability primitives (metrics, logs, traces, Kubernetes) rather than shell access. In the treatment condition, each agent received the Causely MCP server in addition to its baseline toolset; no other changes were made to prompts, models, or tool permissions.

\begin{table}[H]
\centering
\caption{Agent configurations. Baseline tools represent each agent's default toolset; treatment tools are the baseline set extended with the Causely MCP server.}
\label{tab:agents}
\small
\renewcommand{\arraystretch}{1.15}
\footnotesize
\begin{tabular}{p{2.2cm} p{2.2cm} p{4.0cm} p{2.4cm} p{2.6cm}}
\toprule
\textbf{Archetype} & \textbf{Framework} & \textbf{Model} & \textbf{Baseline tools} & \textbf{Treatment tools} \\
\midrule
Productivity AI & Claude Code & \texttt{claude sonnet}           & Shell, \texttt{kubectl} & Baseline + Causely MCP \\
Productivity AI & Codex       & \texttt{gpt-5.4-mini}            & Shell, \texttt{kubectl} & Baseline + Causely MCP \\
Ops AI          & HolmesGPT   & \texttt{gemini-pro-3-flash-lite} & Standard MCP set        & Baseline + Causely MCP \\
Ops AI          & HolmesGPT   & \texttt{claude sonnet}           & Standard MCP set        & Baseline + Causely MCP \\
\bottomrule
\end{tabular}
\end{table}

\subsection{Fault Injection}

The active-fault scenario was realized by injecting a code-level defect into the payment service such that every submitted transaction was rejected. The defect is representative of a common class of application-layer regressions that surface only through their downstream effects: it produces no infrastructure alarms, no resource exhaustion, and no process-level failures, yet it propagates through the transaction path to degrade three downstream services (checkout, accounting, shipping), with elevated error rates and user-visible impact. The ground-truth root cause and blast radius were fixed by construction and verified before each run. The healthy-baseline scenario executed the same application build with the defect absent. Operating both scenarios against an identical topology, traffic profile, and observability configuration isolates the effect of the fault itself.

\subsection{Correctness Rubric}

Correctness was evaluated against a query-type-specific rubric defined prior to data collection. For each query type, the rubric states a sufficient condition for a correct response in both scenarios; responses were scored independently against this rubric and recorded as a binary pass/fail outcome.

\begin{table}[H]
\centering
\caption{Correctness rubric. Queries are grouped by the reliability use case they exercise; the rubric states the condition under which a response is marked correct in each scenario.}
\label{tab:rubric}
\small
\renewcommand{\arraystretch}{1.15}
\begin{tabular}{p{1.1cm} p{3.2cm} p{5.3cm} p{4.6cm}}
\toprule
\textbf{Query} & \textbf{Use case} & \textbf{Active Fault} & \textbf{Healthy Baseline} \\
\midrule
Q1        & Health assessment       & Identifies an active incident in the namespace. & Reports no active incidents. \\
Q2, Q6    & Impact analysis         & Enumerates all three affected downstream services without introducing false positives. & Reports no impacted services. \\
Q3, Q4    & Root cause diagnosis    & Identifies the payment-service defect as the underlying cause (Q4 is the ownership projection: attributes responsibility to the payment service). & Reports no root cause. \\
Q5        & Remediation/triage      & Correctly assesses the safety of the proposed mitigation given the diagnosed root cause. & N/A (query not posed). \\
\bottomrule
\end{tabular}
\end{table}

\subsection{Data Collection}

Every experimental run was instrumented to capture the complete record required for the quantitative results. Persisted artifacts include the query text and the agent's final response, a timestamped execution timeline covering query submission, individual tool invocations, and completion, the input, output, and cached token counts reported by the underlying model, the full sequence of tool calls with their parameters, a USD cost estimate derived from the recorded token counts and the relevant model's published pricing, and the correctness determination obtained by applying the rubric in Table~\ref{tab:rubric}.

\bibliographystyle{plain}
\bibliography{references}

@misc{mckinsey2025stateofai,
  title={The State of AI in 2025: Agents, Innovation, and Transformation},
  author={{McKinsey \& Company}},
  institution={McKinsey \& Company},
  year={2025},
  url={https://www.mckinsey.com/capabilities/quantumblack/our-insights/the-state-of-ai}
}

@misc{mckinsey2025workplace,
  title={Superagency in the Workplace: Empowering People to Unlock AI's Full Potential},
  author={{McKinsey \& Company}},
  institution={McKinsey \& Company},
  year={2025},
  url={https://www.mckinsey.com/capabilities/tech-and-ai/our-insights/superagency-in-the-workplace-empowering-people-to-unlock-ais-full-potential-at-work}
}

@book{beyer2016sre,
  title={Site Reliability Engineering: How Google Runs Production Systems},
  author={Beyer, Betsy and Jones, Chris and Petoff, Jennifer and Murphy, Niall Richard},
  publisher={O'Reilly Media},
  year={2016},
  isbn={978-1491929124}
}

@article{liu2024lostmiddle,
  title={Lost in the Middle: How Language Models Use Long Contexts},
  author={Liu, Nelson F. and Lin, Kevin and Hewitt, John and Paranjape, Ashwin and Bevilacqua, Michele and Petroni, Fabio and Liang, Percy},
  journal={Transactions of the Association for Computational Linguistics},
  volume={12},
  pages={157--173},
  year={2024},
  url={https://arxiv.org/abs/2307.03172}
}

@article{goyal2025contextlength,
  title={Context Length Alone Hurts LLM Performance Despite Perfect Retrieval},
  author={Goyal, Sachin and others},
  journal={arXiv preprint arXiv:2510.05381},
  year={2025},
  url={https://arxiv.org/abs/2510.05381}
}

@misc{anthropic2025pricing,
  title={Plans and Pricing for Claude (Enterprise, Team, Pro)},
  author={{Anthropic}},
  year={2025},
  howpublished={\url{https://www.anthropic.com/pricing}},
  note={Accessed 2025-11}
}

@article{fortune2025openai,
  title={OpenAI Says It Plans to Report Stunning Annual Losses Through 2028 and Then Turn Wildly Profitable Just Two Years Later},
  author={{Fortune}},
  journal={Fortune},
  year={2025},
  url={https://fortune.com/2025/11/12/openai-cash-burn-rate-annual-losses-2028-profitable-2030-financial-documents/}
}

@article{theinformation2025anthropic,
  title={Anthropic Changes Pricing to Bill Firms Based on AI Use Amid Compute Crunch},
  author={{The Information}},
  journal={The Information},
  year={2025},
  url={https://www.theinformation.com/articles/anthropic-changes-pricing-bill-firms-based-ai-use-amid-compute-crunch}
}

@misc{causelydocs,
  title={Causely Technical Documentation: Causal Model and Topology},
  author={{Causely}},
  year={2025},
  howpublished={\url{https://docs.causely.ai}}
}

@misc{gartner2024srehicycle,
  title={Hype Cycle for Site Reliability Engineering, 2024},
  author={{Gartner}},
  institution={Gartner},
  year={2024},
  url={https://www.gartner.com/en/documents/5522895}
}

@misc{gartner2025aisre,
  title={Innovation Insight: AI-Augmented SRE, 2025},
  author={{Gartner}},
  institution={Gartner},
  year={2025},
  url={https://www.bigpanda.io/ar-gartner-ai-augmented-sre/}
}

@article{devops2024continuous,
  title={Continuous Reliability: How QA and SREs Can Improve Their CI/CD Workflow},
  author={DevOps.com},
  journal={DevOps.com},
  year={2024},
  url={https://devops.com/continuous-reliability-how-qa-and-sres-can-improve-their-ci-cd-workflow/}
}

@article{datadog2024feedback,
  title={How a Feedback Loop Improves the Software Development Life Cycle},
  author={Datadog},
  journal={Datadog Blog},
  year={2024},
  url={https://www.datadoghq.com/blog/feedback-loops-progressive-delivery/}
}

@article{computerworld2024observability,
  title={Creating a continuous cycle of improvement with observability},
  author={Computerworld},
  journal={Computerworld},
  year={2024},
  url={https://www.computerworld.com/article/1616457/creating-a-continuous-cycle-of-improvement-with-observability.html}
}

@misc{otel-demo,
  title={OpenTelemetry Demo: A microservices-based distributed system intended to illustrate the implementation of OpenTelemetry},
  author={{OpenTelemetry}},
  year={2024},
  url={https://github.com/open-telemetry/opentelemetry-demo}
}

\end{document}